# A Novel Incremental Clustering Technique with Concept Drift Detection

Mitchell D. Woodbright
School of Computing and Mathematics
Charles Sturt University
Panorama Avenue, Bathurst 2795,
Australia
mwoodbright@csu.edu.au

Md Anisur Rahman
School of Computing and Mathematics
Charles Sturt University
Panorama Avenue, Bathurst 2795,
Australia
arahman@csu.edu.au

Md Zahidul Islam
School of Computing and Mathematics
Charles Sturt University
Panorama Avenue, Bathurst 2795,
Australia
zislam@csu.edu.au

*Abstract*— Data are being collected from various aspects of life. These data can often arrive in chunks/batches. Traditional static clustering algorithms are not suitable for dynamic datasets, i.e., when data arrive in streams of chunks/batches. If we apply a conventional clustering technique over the combined dataset, then every time a new batch of data comes, the process can be slow and wasteful. Moreover, it can be challenging to store the combined dataset in memory due to its ever-increasing size. As a result, various incremental clustering techniques have been proposed. These techniques need to efficiently update the current clustering result whenever a new batch arrives, to adapt the current clustering result/solution with the latest data. These techniques also need the ability to detect concept drifts when the clustering pattern of a new batch is significantly different from older batches. Sometimes, clustering patterns may drift temporarily in a single batch while the next batches do not exhibit the drift. Therefore, incremental clustering techniques need the ability to detect a temporary drift and sustained drift. In this paper, we propose an efficient incremental clustering algorithm called UIClust. It is designed to cluster streams of data chunks, even when there are temporary or sustained concept drifts. We evaluate the performance of UIClust by comparing it with a recently published, high-quality incremental clustering algorithm. We use real and synthetic datasets. We compare the results by using well-known clustering evaluation criteria: entropy, sum of squared errors (SSE), and execution time. Our results show that UIClust outperforms the existing technique in all our experiments.

## I. Introduction

Clustering is a well-known and important data mining process that takes the records of a dataset and produces a set of disjointed groups of records, referred to as clusters so that the records in each cluster are maximally similar. In contrast, their similarity to records in other clusters is minimized [1]-[3]. Traditional static clustering algorithms (such as k-means) are designed to cluster datasets where all the records of the datasets are known in advance, the records within that dataset are static, and the properties of the dataset are never changed [1],[4]-[5]. Due to advances in data processing and gathering technologies, such as data mining and sensors [6], there has been an increase in the amount of data that has been produced in almost all aspects of life [5]. In this scenario, data can have ever-increasing volume and dynamic nature [7]-[8], often arriving in a stream of chunks. Which has led to the development of incremental clustering algorithms. Incremental clustering algorithms have a wide variety of real-world applications, including medical decision systems [9], energy management [10], and credit card fraud detection [11]-[12].

In recent years, incremental clustering has gained a lot of attention, and many algorithms have been developed to try to address the need to cluster data streams [5]-[7], [9]-[10], [13]-[16]. For an incremental clustering algorithm to be suitable for clustering data streams of chunks, the algorithm needs to be able to retain any relevant information learned in the previous clustering results while simultaneously using this information to assist in the clustering process of the current data chunk [9]. We discover two common problems that current incremental clustering algorithms face as follows:

1. The algorithms need to be able to reduce the memory overhead of clustering large quantities of data [14], as it can be difficult to store all records received from a data stream [6],[12]. This is attributed to data streams being large [8],[13].

2. The algorithms need to be able to adjust for temporary and sustaining changes that can occur within a data stream, commonly referred to as concept drift [3],[7]-[8],[13]. If a concept drift is not appropriately addressed, it can cause a reduction in the algorithm's performance and accuracy [8].

Therefore, in this research, we propose a novel incremental clustering algorithm called UIClust (Unsupervised Incremental Clustering Algorithm) that addresses the above issues. UIClust reduces its memory overhead of clustering a data stream of chunks by achieving compactness of representation, which reduces clusters down into key pieces of information for processing. This allows the algorithm to discard records once they have been learned. It also contains a unique parallel clustering component along with a concept drift detection component to handle potential concept drifts. In this paper, by parallel clustering, we mean multiple clustering being performed separately at a given point of time during the incremental learning, and we do not mean the parallel processing of a clustering operation.

The parallel clustering component uses a three-strike policy to distinguish between temporary and sustaining concept drifts. This unique parallel clustering component is our main contribution in this paper. We compare the performance of our proposed technique with a modern incremental clustering algorithm proposed by Zheng, Huo, & Fang [16], termed as Supervised Adaptive Incremental Clustering (SAIC) using well-known clustering evaluation criteria including entropy, sum of squared errors (SSE), and execution time.

The organization of the rest of this paper is as follows: Section II is a literature review on traditional and incremental clustering, Section III discusses our proposed algorithm which is broken into components and explained in detail. In section IV, we present our evaluation of UIClust compared to a modern incremental clustering algorithm called SAIC. Section VI is the last; here, we will be giving our conclusion, followed by a quick mention of future work.

## II. RELATED WORK WITH PROBLEM DISCUSSION

K-means is a simple and popular traditional static clustering algorithm [17]. K-means requires all the records in advance before the algorithm starts. This is a common theme amongst traditional clustering algorithms. For example, DSCAN [18], GenClust [19], GenClust++ [20], OPTICS [21], and HeMI [22] all require the records in advance. Therefore, incremental clustering algorithms are designed to overcome this issue.

Zheng, Huo, Guo & Fang [16] propose a supervised incremental clustering algorithm for clustering a data stream of chunks. Their algorithm does not address the issue of concept drifts. Therefore, whenever a concept drift occurs, this can cause a degradation in SAIC's current model, causing inaccurate results. This is because "a set of examples have legitimate class labels at one time and has different legitimate labels at another time" [11]. SAIC's lack of concept drift handling causes it to bring previous clusters into the current data chunk that are no longer valid.

Li, Li, Wang & Zhai [12] propose an integrated clustering framework for clustering categorical data streams with the presence of concept drift. In their method, whenever they detect a concept drift, the algorithm simply retrains the model with the current data chunk (which is not always ideal). Through our research, we have observed that whenever an algorithm with a concept drift detection mechanism discovers a concept drift, it merely instructs the algorithm to retrain the model [12]. This doesn't allow the algorithm to distinguish between temporary (caused by corrupted data chunks, etc.) and sustaining concept drifts. Therefore, we believe our proposed method must use a parallel-clustering component to help distinguish between temporary concept drifts and sustaining concept drifts to give better high-quality results under concept drift scenarios.

## III. UICLUST: A NOVEL INCREMENTAL CLUSTERING TECHNIQUE

In this paper, we propose a novel incremental clustering algorithm called UIClust. The aim of designing the algorithm is to solve two issues discussed in the introduction and to produce a more effective and efficient incremental clustering algorithm in comparison to a modern algorithm. UIClust uses compactness of representation to address the need to retain all records within a data stream of chunks. Instead, it only keeps the key information necessary to assist in incremental clustering (discarding all learned records). UIClust implements a concept drift detection component along with a parallel clustering component to monitor and adjust for changes that can occur in data stream clustering (including concept drifts). The parallel clustering component activates whenever a concept drift occurs, allowing the main clustering results to have three consecutive chances to stabilize (become non-concept-drift) before requiring exchanging the current model and results for the one created by the parallel clustering. However, unlike other concept drift detection methods that retrain from scratch as soon as a potential concept drift has been detected [8],[12], our algorithm uses the parallel clustering component to avoid hastily retraining the model in case of occurrences of temporary (non-sustaining) concept drifts that can be caused by faulty sensors, etc. This makes our method more robust to the occurrences of temporary concept drifts. Additionally, the parallel clustering component gives the user better-quality replacement for the current model, because our method already has up to three data chunks worth of new knowledge when it does detect the requirement for retraining, instead of having to restart the algorithm and learn from scratch from the current data chunk.

### A. Notation and Definitions

To discuss the main components of our proposed method, we shall first introduce basic notations and definitions. We present a data stream of chunks as a sequence $S^t, S^{t+1}, S^{t+2}, \ldots, S^n$ of $n$ data chunks. Where the superscript $t$ is the integer timestamp of the data chunk's arrival. Each record $R_m$ within a data chunk $S^t$ contains $n$ attributes $A_1^t, A_2^t, \ldots, A_n^t$.

UIClust produces two types of results, that is, it contains the main clustering result $C^t$ and a parallel clustering result $P^c$ (in the presence of concept drifts). The parallel clustering result's superscript $c$ is an integer counter to track the number of parallel clustering results. Each clustering result $C^t$ contains a set of clusters $c_1^t, c_2^t, \ldots, c_k^t$ and an outlier counter value $outliers^t$. All clusters within a clustering result contain four key pieces of information $c_k^t = \{x_k^t, \lambda_k^t, \psi_k^t, \Delta_k^t\}$. $x_k^t$ is a set of $n$ centroid values of the cluster, which is the average attribute values of all records that have been assigned to the cluster. The maximum distance threshold $\lambda_k^t$ is a radius value used in the clustering process to determine if a record is added to that cluster or not. $\psi_k^t$ is a learning rate, which is an integer value that contains the number of records that have been assigned to the cluster over the entire period the cluster has existed, not just for that timestamp. $\Delta_k^t$ is an integer value of the current number of records assigned to that cluster for that timestamp, this is used in the concept drift detection component.

UIClust uses a user-defined maximum cluster dissimilarity threshold "$dThresh$" and outlier ratio threshold "$oThresh$" in detecting concept drifts. "$dThresh$" is a user-defined numerical threshold value, used to compare the dissimilarity between two clustering observations in the concept drift detection method. And "$oThresh$" is a user-defined numerical value to monitor the number of outliers in an observation to detect potential concept drifts.

### B. Basic Components

In this section, we first present the four basic components used in UIClust and then move onto a quick summary of how our proposed algorithm works. The main components are as follows:

#### 1) Component 1: Initial Clustering with K-means

This component is used to discover clusters within a data chunk whenever the model needs to be either initially trained or retrained (Component 4) and summarize the cluster's results into only the key pieces of information required to cluster the results incrementally. This component begins by using k-means (with a user-defined $k$ value) to discover the $k$ number of clusters. Then for each cluster $c_k^t$ within the clustering results $C^t$, it summarizes the clusters into four key pieces of information $c_k^t = \{x_k^t, \lambda_k^t, \psi_k^t, \Delta_k^t\}$. The centroid values $x_k^t$ of the cluster $c_k^t$ is calculated by k-means. The maximum distance threshold $\lambda_k^t$ is calculated by $getMaxDist(c_k^t)$ method. The $getMaxDist(c_k^t)$ method calculates the Euclidean Distance between the cluster's centroid $x_k^t$ and farthest Euclidean Distance record $R_m$ assigned to that cluster.

The learning rate $\psi_k^t$ and the number of assigned records $\Delta_k^t$ are initialized as the number of records assigned to that cluster. Finally, all pieces of information are assigned to the cluster's results by the $addResults(\lambda_k^t, \Delta_k^t, \psi_k^t)$ method and all earned records are forgotten. This component is summarized in the following figure (Fig.1. Algorithm 1: Initial clustering with k-means).

---
**Algorithm 1:** Initial clustering with k-means
**Input:** A data chunk $S^t$, $k$ number of clusters
**Output:** Clustering result $C^t = \{c_1^t, c_2^t..., c_k^t\}$
**begin**
    $C^t \Leftarrow kmeans(S^t, k)$;
    **for** ($c_k^t$ **in** $C^t$) **do**
        $\lambda_k^t \Leftarrow getMaxDist(c_k^t)$;
        $\psi_k^t \Leftarrow |c_k^t|$;
        $\Delta_k^t \Leftarrow |c_k^t|$;
        $c_k^t \Leftarrow addResult(\lambda_k^t, \Delta_k^t, \psi_k^t)$;
    **end**
    output $C^t$;
**end**

---

Fig. 1. Algorithm 1: Initial clustering with k-means

*2) Component 2: DistClust Incremental Clustering*

This component is the incremental clustering component that we have named DistClust. It uses the current data chunk $S^t$ and the previous clustering results $C^{t-1}$ to incrementally update the previous clustering result with the new records from the current data chunk to produce the current clustering results $C^t$.

The DistClust component starts by getting all the maximum distance thresholds $\lambda^t$, using the $getAllMaxDistThresholds(\lambda^t)$ method. Then for each record within the data chunk, DistClust calls the $getClostestClustIndex(R_m, C^{t-1})$ method to retrieve the index $k$ of the closest (Euclidean Distance) cluster to the new record. Then the distance between the record and the nearest cluster is calculated using the $getDistance(R_m, c_k^{t-1})$ method. If the distance is less than or equal to the cluster's maximum distance threshold ($dis \leq \lambda_k^t$), then the cluster is updated. Else, the number of outliers "$outliers^t$" discovered is incremented by 1. This component is displayed in more detail in Fig.2, where we presented the DistClust algorithm.

Whenever a new record is assigned to a cluster, the learning rate $\psi_k^t$ and the number of assigned records $\Delta_k^t$ are incremented by 1. Then each centroid value $(x_1^t, x_2^t, ..., x_n^t)$ are updated using the following equation.

$$new\ x_i^t = \left(1 - \frac{1}{\psi_k^t}\right) * x_i^t + \frac{1}{\psi_k^t} * A_i^t \quad (1)$$

Where $x_i^t$ is the $i$th centroid value of the cluster. $\psi_k^t$ is the updated learning rate and $A_i^t$ is the $i$th attribute value of the newly assigned record $R_m$. This equation is similar to the one we learned from Zheng, Huo, and Fang [16]. This component is summarized in the following figure (Fig.2. Algorithm 2: DistClust).

---
**Algorithm 2:** DistClust
**Input:** A data chunk $S^t$, Previous clustering results $C^{t-1}$
**Output:** Clustering result $C^t = \{c_1^t, ..., c_k^t, outliers^t\}$
**Initialization:**
$k = 0$; /* $k$ is the index of a cluster */
**begin**
    $\lambda^t \Leftarrow getAllMaxDistThresholds(\mathbf{C}^{t-1})$;
    **for** ($m = 1$ **to** $|S^t|$) **do**
        $k \Leftarrow getClosestClustIndex(R_m, \mathbf{C}^{t-1})$;
        $dist \Leftarrow getDistance(R_m, \mathbf{C}_k^{t-1})$;
        **if** ($dist \leq \lambda_k^t$) **then**
            $C^t \Leftarrow updateCluster(R_m, \mathbf{C}^t, \mathbf{C}^{t-1})$;
        **end**
        **else**
            $C^t \Leftarrow updateOutliers(R_m)$;
        **end**
    **end**
    output $C^t$;
**end**

---

Fig. 2. Algorithm 2: DistClust

*3) Component 3: Concept Drift Detection*

This component is our concept drift detection method. It requires the current clustering result $C^t$, the previous clustering result $C^{t-1}$, a user-defined outlier threshold "$oThresh$" and a user-defined maximum cluster dissimilarity threshold "$dThresh$" to detect potential concept drifts. Here we consider two cases from which a concept drift can occur (which are very similar to the cases that we learned from [12]) as follows:

1. Case 1: If the number of outliers "$outliers^t$" exceed a predetermined threshold "$oThresh$". So, we must consider new clusters that may have emerged, or a change has occurred in the stream.

2. Case 2: If the number of outliers "$outliers^t$" is within the predetermined outlier threshold "$oThresh$," but there has been a significant change in the distribution of records assigned to a cluster in the current clustering results compared to the previous clustering results $C^{t-1}$. Indicating that a change has occurred in the stream.

The concept drift detection method starts by checking if the ratio of outliers is within the user-defined outlier threshold value "$oThresh$." If the ratio of outliers exceeds this threshold value, the component returns a potential concept drift. If not, for each cluster the component calls the $getEachClusterDistrib(c_k^t)$ method to get the distribution of records assigned to the cluster for the current results $C^t$ and previous $C^{t-1}$. The change in distribution and percentage of change is then calculated. Finally, if the percentage of change is not within the current maximum dissimilarity threshold, then a concept drift is flagged. Else, the component is finished, and it returns that no concept drifts have occurred. This component is displayed in more detail in Fig.3. Algorithm 3: Concept Drift Detection.

*4) Component 4: Parallel Clustering*

The parallel clustering component is used to handle whenever a potential concept drift has occurred. It is designed for handling temporary and sustaining concept drifts. This component is activated whenever a concept drift has occurred. It only deactivates once the main clustering results stabilized (become non-concept-drift) or three consequential parallel.

```
Algorithm 3: Concept Drift Detection
Input: Current clustering result C^t, Previous clustering result C^{t-1},
       threshold, dThresh a max dissimilarity threshold
Output: Boolean value of isConceptDrift
Initialization:
isConceptDrift = false;
changeDistrib = 0;
percentChange = 0;
begin
    if (outlier^t > oThresh) then
    |   isConceptDrift = true;
    end
    else
    |   for (k = 1 to |C^t| - 1) do
    |       Δ_k^t ← getEachClusterDistrib(C_k^t);
    |       Δ_k^{t-1} ← getEachClusterDistrib(C_k^{t-1});
    |       changeDistrib = Δ_k^t − Δ_k^{t-1};
    |       percentChange = changeDistrib ÷ c_k^{t-1};
    |       if (percentChange > dThresh) then
    |       |   isConceptDrift = true;
    |       |   break for loop;
    |       end
    |   end
    end
    return isConceptDrift;
end
```

Fig. 3. Algorithm 3: Concept Drift Detection

clustering results have been created. The parallel clustering component can be reduced to the following three steps:

1. Step 1: In the case of a suspected concept drift being detected from the concept drift detection method (Component 3), the parallel clustering component is activated, and the initial parallel clustering result $P^c$ is created (where $c = 1$) by performing k-means (Component 1) on the current data chunk $S^t$. At the end of this step, the parallel counter $c$ is incremented by 1 and the parallel clustering result $P^c$ is returned. The component then waits for the next main clustering results to move onto step 2.

2. Step 2: While the main clustering results have not stabilized, or the parallel clustering counter $c$ has not reached 3. The previous parallel clustering results $P^{c-1}$ is incrementally updated with the new data chunk using distClust (Component 2) to produce the latest parallel clustering result $P^c$. The current results are then checked for potential concept drifts (Component 3). If a concept drift occurs within the parallel clustering results, the parallel clustering results are disregarded, and the parallel clustering retrains using k-means (Component 1) and the current data chunk $S^t$. If the main clustering results have stabilized or the parallel clustering results have had 3 then the component moves onto step 3. Else, step 2 is repeated with a new data chunk.

3. Step 3: If the main clustering results have stabilized, then the parallel clustering results are disregarded, the parallel clustering counter $c$ is reset to 1, and the component is deactivated. If not, the parallel clustering results replace the main clustering results, and the main clustering results are disregarded. The parallel clustering counter $c$ is reset to 1, and the component is deactivated.

For illustrative purposes, we have created two figures to assist with the comprehension of this component (Fig.4 & Fig.5). The diagram components are as follows:

- The dotted lines indicate time. For example, the horizontal lines indicate the direction of flow of the method, and the vertical lines indicate different timestamp periods.
- The top black circles indicate the input data chunks and are filled with green dots representing records.
- The circles with blue dots indicate a clustering result, where the blue dots are the centroids. If the circles containing the blue dots are red, this shows a concept drift.
- The rectangle with UIClust is a black box representation of the algorithm working.
- The three black dogs in Fig.5 is to skip a data chunk (for illustrative purposes only).

We will now give a quick overview of each diagram for clarity.

The first diagram (Fig.4) shows the occurrence of a temporary drift. The temporary drift occurs at timestamp $S^t$. It is observed that the concept drift has happened due to a change in distribution from the previous results with the current data chunk. Therefore, a parallel clustering result has been initiated. However, at timestamp $S^{t+1}$ the distribution returns to normal, and the parallel component is disabled and forgotten.

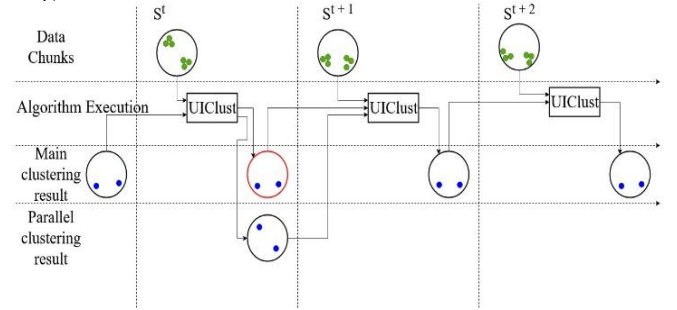

Fig.4. Temporary concept drift

The second diagram (Fig.5) shows the occurrence of a sustaining drift. This drift occurs at $S^t$ and sustains for the remainder data chunks. It is observed that after three parallel clustering results the concept drift is sustaining, and the main clustering fails to stabilize. Therefore, at time $S^{t+3}$ the parallel clustering results become the new main results.

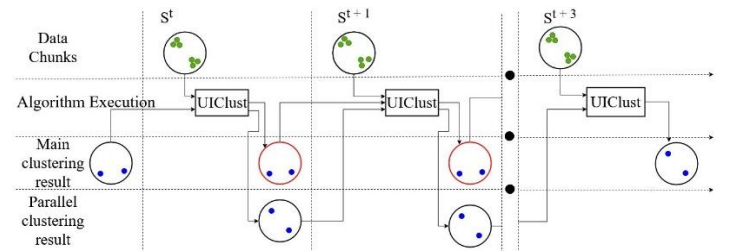

Fig.5. Sustaining concept drift

### C. Proposed UIClust Technique

Now we explain the UIClust technique using the components that we have described previously. For the first data chunk in the stream, the algorithm initializes a set of clusters using the k-means component. This is then summarized into the key pieces of information for later processing. Then for every sequential data chunk, it uses DistClust to update the clustering results with the new data provided. It then checks if a concept drift has occurred. If a concept drift has occurred, then the parallel clustering component is activated, giving the main results three consecutive data chunks to stabilize. However, if it doesn't stabilize after three consecutive data chunks, the main clustering results are replaced with the parallel clustering results, and the algorithm returns to incrementally clustering on the next data chunk as usual. However, if the main stabilizes at any point while the parallel clustering component is activated, the parallel clustering component is disabled, and parallel clustering results are forgotten. For more details, please see Fig.6, where we have presented the UIClust algorithm.

```
Algorithm 4: UIClust
Input: A data chunk S¹, k number of cluster, oThresh an outlier threshold, dThresh a
       max dissimilarity threshold
Output: A set of Clusters C^t
Initialization: newBatch = true, t = 1; /* t is the time stamp */, c = 1; /* parallel counter */,
isConceptDrift = false, isParaClust = false isMainClust = false
while (newBatch == true) do
    if (t == 1 & isConceptDrift == false) then
        C^t ⇐ kmeans(S^t, k);
    end
    else if (t != 1 & isConceptDrift == false) then
        C^t ⇐ distClust(S^t, C^{t-1});
        isConceptDrift ⇐ conceptDriftDetection(C^t, C^{t-1}, oThresh, dThresh);
        if (isConceptDrift == true) then
            P^c ⇐ kmeans(S^t, k);
            c = c + 1;
        end
    end
    else if (t != 1 & isConceptDrift == true) then
        if (c ≤ 3) then
            C^t ⇐ distClust(S^t, C^{t-1});
            isMainClust ⇐ conceptDriftDetection(C^t, C^{t-1}, oThresh, dThresh);
            if (isMainClust == false) then
                c = 1;
                isConceptDrift = false;
            end
            else
                P^c ⇐ distClust(S^t, P^{c-1});
                isParaClust ⇐ conceptDriftDetection(P^c, P^{c-1}, oThresh, dThresh);
                if isParaClust == true then
                    P^c ⇐ kmeans(S^t, k);
                end
                c = c + 1;
            end
        end
        if (c == 4) then
            C^t = P^{c-1};
            c = 1;
            isConceptDrift = false;
            isMainClust = false;
        end
    end
    t = t + 1;
    if (isConceptDrift == true & isMainClust == true & isParaClust == false) then
        C^t = P^{c-1};
    end
    newBatch ⇐ doYouWishToContinue();
    if (newBatch != true) then
        output C^t;
        break;
    end
    S^t ⇐ nextDataChunk();
    output C^t;
end
```

Fig.6. Algorithm 4: UIClust

IV. EXPERIMENTAL RESULTS AND DISCUSSION

This section first defines some of the evaluation criteria used in our experiments and then discusses the experimental setup. We then discuss our experiments conducted with synthetic datasets and finishing with a final discussion on our experiments that uses UCI real-world datasets.

A. *Definitions of the Cluster Evaluation Criteria*

In our experiments, we used the following evaluation criteria: Entropy, Sum of Squared Errors, True Cluster Value, and Runtime.

In our synthetic dataset experiment Scenario 1, we introduce a cluster evaluation measure we call True Cluster Value (TCV). The TCV is calculated by converging all data chunks into one large dataset and calculating the centroid values of all clusters. These centroid values are considered the TCV of each cluster. Then to evaluate the algorithm, we compare the centroids discovered at the end of the data stream to the corresponding TCV. Here the number closest to 0 indicates a better result.

B. *Experimental Setup*

For all experiments, we use an internal switch statement to set UIClust's $k$ number of clusters value to the number of unique cluster labels for each inputted data chunk. Also, the outlier threshold "$oThresh$" and maximum dissimilarity value "$dThresh$" was set to 0.18 and 0.6 respectively for all synthetic datasets, and 0.18 and 0.4 on the real-world UCI datasets. We chose these values for UIClust as these showed the optimal results on our preliminary tests of each type of dataset. It was observed that if the "$oThresh$" ratio is set too high, it causes the algorithm to overlook concept drifts. In contrast, when this ratio is set too low, the algorithm becomes too sensitive to outliers, leading to false-positive concept drift detections and redundant parallel clustering activation. Furthermore, if the "$dThresh$" value is chosen incorrectly, it can cause false positives concept drift detection (when the value is set too small) and false negatives (when the value is too large). Incorrectly setting the "$dThresh$" value can also cause redundant parallel component activation.

For SAIC, we set the maximum number of iterations value "MaxIter" to 20 to coincide with their experiments [16], and we also set the minimum number of data points in a cluster value "θ" to 2. These values gave SAIC the best results on all our preliminary tests.

It is essential to mention that for all experiments, the datasets are preprocessed using min-max normalization to reduce the impact of attributes with larger scales of numerical values before being processed into data streams of chunks (discussed later).

C. *Data Stream of Chunks Setup*

To generate a data stream of chunks, we take an entire dataset $D$ and split it into $n$ number of data chunks $S^1, S^2, ... S^n$. We will now give an example of this using a toy dataset. Here we have a toy dataset (Table I) containing 8 records $R_1, R_2, ..., R_8$, with each record containing 2 attributes $A_1, A_2$ and a class label.

TABLE I.   TOY DATASET

| Record | $A_1$ | $A_2$ | Class Label |
|---|---|---|---|
| $R_1$ | 0.052 | 0.153 | 1 |
| $R_2$ | 0.061 | 0.252 | 1 |
| $R_3$ | 0.046 | 0.175 | 1 |
| $R_4$ | 0.055 | 0.183 | 1 |
| $R_5$ | 0.957 | 0.858 | 2 |
| $R_6$ | 0.965 | 0.752 | 2 |
| $R_7$ | 0.957 | 0.858 | 2 |
| $R_8$ | 0.965 | 0.752 | 2 |

In this example, our dataset is to be divided into two separate data chunks. Whenever a dataset is divided, we take the $n$ records from each unique class to ensure we captured the nature of each class. In our example, the toy dataset needs to be divided into two equal-size data chunks containing 4 records each. Therefore, we start by taking the first two records from each class.

### D. Experimental Results and Discussion on Synthetic Data

In our synthetic data experiments, we create four synthetic datasets to generate the four data streams of chunks for our experiments. Each chunk contains 150 records, and whenever a concept drift occurs, labels are reassigned randomly. For example, if a data chunk contains 2 unique class labels and a concept drift occurs. The records containing the class label "1" will now be assigned to class label "2" and vice versa. We will now discuss the four different generated synthetic data streams.

#### a) Synthetic Data Stream of Chunks Setup

The first data stream we generated (SDWCD) contains 1 temporary drift that occurs that is shortly followed by a sustaining concept drift. That is for the first 2 data chunk ($t = 1, 2$) the number of clusters is 5, and they all contain 30 records each, then at $t = 3$ when the number of clusters is reduced to 1, a temporary concept drift occurs which only changes the class labels for that data chunk. Now for $t = 4$ and 5, the number of clusters and labels go back to the same logic that is in $t = 1$ and 2. Then at $t = 6$, a sustaining concept drift occurs, which sustains for the remainder of the stream ($t = 6$ to 10). In the sustaining concept drift data chunks, the number of clusters is equal to 3. This data stream is summarized in Table II.

TABLE II. SYNTHETIC DATA STREAM OF CHUNKS WITH CONCEPT DRIFT (SDWCD)

| $t =$ | 1 | 2 | 3 | 4 | 5 | 6 | 7 | 8 | 9 | 10 |
|---|---|---|---|---|---|---|---|---|---|---|
| $|c_k^t|$ | 30 | 30 | 150 | 30 | 30 | 50 | 50 | 50 | 50 | 50 |
| $|C^t|$ | 5 | 5 | 1 | 5 | 5 | 3 | 3 | 3 | 3 | 3 |

For the second data stream (SDCCL), the first four data chunks contain the same number of clusters (5), and then at $t = 5$ the number decreases to 1 just for this timestamp. However, the cluster's location has moved. This little variance in cluster location is then continued for the next two data chunks. At timestamp $t = 6$ and 7, the distribution returns to 5 clusters, and the location of the other 4 returning clusters are slightly moved as well. This data stream is summarized in Table III.

TABLE III. SYNTHETIC DATA STREAM WITH A CHANGE IN CLUSTER LOCATION (SDCCL)

| $t =$ | 1 | 2 | 3 | 4 | 5 | 6 | 7 |
|---|---|---|---|---|---|---|---|
| $|c_k^t|$ | 30 | 30 | 30 | 30 | 150 | 30 | 30 |
| $|C^t|$ | 5 | 5 | 5 | 5 | 1 | 5 | 5 |

The third data stream (100 Data Chunks NCD), where NCD stands for "no concept drift"; this data stream contains 100 data chunks, which all contain 5 clusters, all of which contain 30 records each. This data stream does not shift cluster locations or change labels. We have chosen not to include this dataset in a table due to its simplicity.

The fourth and last synthetic data stream (1000 Data Chunks WCD), where WCD stands for "With Concept Drift," contains 1000 data chunks that change both the number of clusters every 100 data chunks and the label distribution (sustaining concept drifts). This data stream was created to emulate a large data stream that contains several changes to its logic. This is summarized in Table IV.

#### b) Synthetic Data Experimental Results

In this section, we compare SAIC and UIClust in three different scenarios.

TABLE IV. SYNTHETIC DATA STREAM OF CHUNKS CONTAINING 1000 DATA CHUNKS

| $t =$ | 1-100 | 101-200 | 201-300 | 301-400 | 401-500 | 501–600 | 601–700 | 701–800 | 801-900 | 901-1000 |
|---|---|---|---|---|---|---|---|---|---|---|
| $|c_k^t|$ | 30 | 37/38 | 30 | 50 | 30 | 37/38 | 30 | 75 | 50 | 30 |
| $|C^t|$ | 5 | 4 | 5 | 3 | 5 | 4 | 5 | 2 | 3 | 5 |

Scenario 1 is to evaluate how well both algorithms deal with the changes that can occur in clustering a data stream. Here we have used the SDWCD data stream (discussed earlier) to simulate a stream that contains a temporary drift shortly followed by a sustaining drift.

Scenario 2 is to compare SAIC and UIClust's ability to discover the true nature of clusters in the occurrence of a stream that reduces the number of the clusters temporarily and then return with a slight shift in location (SDCCL). That is, we are trying to see if the algorithms can discover the true centroid value (TCV) of each cluster.

Scenario 3 and final scenario is simply a runtime comparison. SAIC's and UIClust's runtime over two synthetic data streams (one with 100 data chunks and no concept drift, while the other has 1000 data chunks and 100 concept drifts) are compared to evaluate how efficient the algorithms are over a more extended stream. We will now discuss the results of each scenario.

SCENARIO 1: ABILITY TO HANDLE TEMPORARY AND SUSTAINING CONCEPT DRIFTS

In this scenario, we used the external cluster evaluation criteria name entropy [21], and the average number of clusters discovered for each data chunk (over 20 runs) is compared to the actual number of clusters (ground truth) in each data chunk of the SDWCD stream. The results of this scenario showed that SAIC achieved a perfect score entropy (0) for every cluster in each data chunk. It was also observed that UIClust performed with the same amount of precision, achieving the same perfect score. Following, we compared the number of clusters discovered in each data chunk by both algorithms with the actual number of clusters (ground truth) in each data chunk in the same SDWCD stream. Here SAIC failed to discover the true number of clusters (on average) for more than half of the data chunks. This was attributed to the changes in cluster labels. The changes are causing SAIC to create incorrect cluster formations. Also, we noticed that SAIC achieved different results for each time we ran the algorithm. Therefore, to make a fair comparison, we decided to run both algorithms over the data stream 20 times and then averaged the number of clusters found. SAIC still failed to discover the correct number of clusters. However, in contrast to SAIC, UIClust always managed to find the correct number of clusters. This test shows that UIClust is very good at handling concept drifts, while SAIC is not. The resulting number of clusters discovered in SDWCD for each data chunk is summarized in Fig.7. Number of clusters discovered in SDWCD by SAIC and UIClust.

SCENARIO 2: CAPTURING CLUSTERS TRUE NATURE

This scenario compares the centroid values of each cluster discovered by SAIC and UIClust at the completion of the SDCCL data stream to the true centroid values (TCV) of the clusters in the stream. In this data stream, there is no label changing concept drift, instead, the number of clusters is changed for 1 data chunk ($t = 5$), and the distribution of cluster locations has been slightly shifted for the last 3 data chunks ($t = 5$ to 7). For this test, we combined all the data chunks into a single dataset and calculated the 5 cluster's TCV. We ran SAIC and UIClust on the synthetic data stream to obtain their centroid values at the end of the stream. Here we observed that when SAIC reduces the number of clusters down to 1 in $t = 5$ it forgets all previous knowledge. However, when UIClust discovers a change in distribution, it's parallel clustering component (Component 4) activates, and it does not lose this information. Instead, the main clustering stabilizes before the end of the stream because the distribution returns to normal, and all clusters are retained. This superiority in UIClust's ability to handle temporary changes in the logic of the data stream has allowed it to capture the TCV of each cluster, while SAIC has not. These results are summarized in the following two tables (Table V & Table VI).

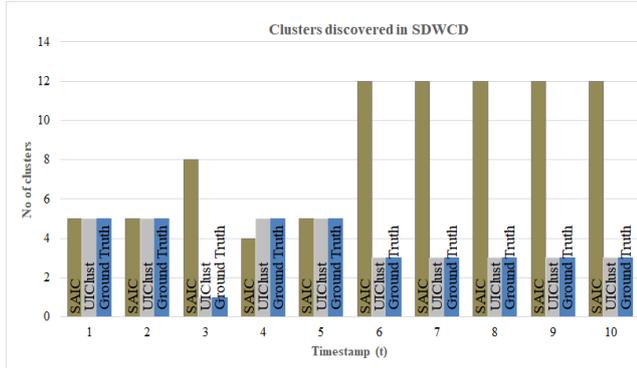

Fig.7. Number of clusters discovered in SDWCD by SAIC and UIClust

TABLE V. TRUE CLUSTER VALUES OF SAIC USING THE SDCCL DATA STREAM

| Cluster Number | TCV $x_1^t$ | TCV $x_2^t$ | SAIC $x_1^t$ | SAIC $x_2^t$ | SAIC Distance |
|---|---|---|---|---|---|
| C1 | 0.117 | 0.884 | 0.151 | 0.857 | 0.044 |
| C2 | 0.885 | 0.885 | 0.852 | 0.860 | 0.041 |
| C3 | 0.527 | 0.635 | 0.528 | 0.637 | **0.000** |
| C4 | 0.117 | 0.111 | 0.160 | 0.145 | 0.054 |
| C5 | 0.877 | 0.117 | 0.841 | 0.153 | 0.051 |

TABLE VI. TRUE CLUSTER VALUES OF UIClust USING THE SDCCL DATA STREAM

| Cluster Number | TCV $x_1^t$ | TCV $x_2^t$ | UIClust $x_1^t$ | UIClust $x_2^t$ | UIClust Distance |
|---|---|---|---|---|---|
| C1 | 0.117 | 0.884 | 0.117 | 0.884 | **0.00** |
| C2 | 0.885 | 0.885 | 0.885 | 0.885 | **0.00** |
| C3 | 0.527 | 0.635 | 0.527 | 0.635 | **0.00** |
| C4 | 0.117 | 0.111 | 0.117 | 0.111 | **0.00** |
| C5 | 0.877 | 0.117 | 0.877 | 0.117 | **0.00** |

SCENARIO 3: RUNTIME OVER LONG STREAMS

In this scenario, we ran SAIC and UIClust over two synthetic data streams, and the runtime was captured. Here we used the synthetic data stream with 100 data chunks with no concept drift and the synthetic data stream with 1000 data chunks with 100 concept drift. In both cases, UIClust outperformed SAIC by at least 24%; this is due to the iterative nature of SAIC to discover the optimal number of clusters and UIClust's reduction of memory overhead. The results are summarized in Table VII. The runtime of SAIC and UIClust in seconds over two large two-dimensional synthetic data streams.

TABLE VII. THE RUNTIME OF SAIC AND UIClust IN SECONDS OVER TWO LARGE TWO-DIMENSIONAL SYNTHETIC DATA STREAMS

| Algorithm runtime (seconds) | 100 Data Chunks (NCD) Average | 100 Data Chunks (NCD) Total | 1000 Data chunks (WCD) Average | 1000 Data chunks (WCD) Total |
|---|---|---|---|---|
| SAIC Mean | 0.004 | 0.423 | 0.003 | 3.165 |
| UIClust Mean | **0.003** | **0.331** | **0.002** | **2.399** |

*E. Experimental Results and Discussion on Real-World Data*

In our real-world data experiments, we have created two data streams of chunks made from the UCI "Wine Quality" dataset and UCI "Pen-Based Recognition of Handwritten Digits DataSet" [23]. These datasets are pre-processed before being converted into a stream of chunks (normalized, etc.) and are summarized in Table VIII.

TABLE VIII. REAL-WORLD DATASETS OVERVIEW AFTER PRE-PROCESSING

| Name of dataset | Total No. of records | No. of numerical attributes | Number of classes |
|---|---|---|---|
| Pen-Based Recognition of Handwritten Digits (Pen-Digits) | 10992 | 16 | 10 |
| Wine Quality | 4898 | 11 | 7 |

For the real-world experiments, we use one test scenario to compare SAIC and UIClust. This was to evaluate both methods' abilities to handle temporary and sustaining changes that could occur in a real-world setting using real-world data. Whenever a concept drift occurs during our experiments, we do not alter the class label. Additionally, for each data stream that we have generated, we created $n$ artificial classes (where $n$ coincides with the number of attributes in the dataset), to help give a fair comparison of both the clustering algorithms.

To generate an artificial class, we use a binning technique to divide the range of its coinciding attribute values into $n$ bins (where $n$ is the number of unique class labels of the dataset). Then we generate an artificial class by taking the coinciding attribute values, checking each record's attribute value, and assigning each record an artificial class value based on each bins' range. For example, here, we have created a toy data chunk that contains 6 records, 4 attributes, and 3 unique classes (see Table IX). Therefore, in this example, we are required to generate 4 artificial classes (AC1,…, AC4) and 3 artificial class values for testing (1,..3).

Knowing the attributes' range values are between 0 and 1 inclusively, and that the number of classes is 3. We generate artificial class values by creating 3 bins. Therefore, the range of artificial class value 1 (Bin 1) would be (0, 0.33), artificial class value 2 (Bin 2) (0.34, 0.66), and artificial class value 3 (Bin 3) (0.67, 1.0). Then using each record's attribute values, we would use this information to assign it's coinciding artificial class value to its artificial class.

TABLE IX. TOY DATA CHUNK WITH FOUR ATTRIBUTES

| Record | $A_1$ | $A_2$ | $A_3$ | $A_4$ | Class Label |
|---|---|---|---|---|---|
| $R_1$ | 0.052 | 0.153 | 0.772 | 0.953 | 1 |
| $R_2$ | 0.061 | 0.252 | 0.761 | 0.952 | 1 |
| $R_3$ | 0.957 | 0.858 | 0.257 | 0.258 | 2 |
| $R_4$ | 0.965 | 0.752 | 0.265 | 0.252 | 2 |
| $R_5$ | 0.543 | 0.533 | 0.012 | 0.092 | 3 |
| $R_6$ | 0.496 | 0.488 | 0.022 | 0.097 | 3 |

For example, record $R_1$ attribute $A_1$ value is 0.052 because this attributes value is between 0 and 0.33, then the record's Artificial Class 1 (AC1) would be set as 1 (Bin 1). $R_1$ attribute $A_2$ value is 0.153 because it is between 0 and 0.33, then the record's Artificial Class 2 would be set as 1. $R_1$ attribute $A_3$ value is 0.772 because it is between 0.67 and 1.0, then Artificial Class 3 would be set as 3. And so on. The results of generating the artificial classes for Table X is displayed in Table IX.

TABLE X. DATA CHUNK WITH ARTIFICIAL CLASSES (AC)

| Record | $A_1$ | $A_2$ | $A_3$ | $A_4$ | AC1 | AC2 | AC3 | AC4 |
|---|---|---|---|---|---|---|---|---|
| $R_1$ | 0.052 | 0.153 | 0.772 | 0.953 | 1 | 1 | 3 | 3 |
| $R_2$ | 0.061 | 0.252 | 0.761 | 0.952 | 1 | 1 | 3 | 3 |
| $R_3$ | 0.957 | 0.858 | 0.257 | 0.258 | 3 | 3 | 2 | 2 |
| $R_4$ | 0.965 | 0.752 | 0.265 | 0.252 | 3 | 3 | 2 | 2 |
| $R_5$ | 0.543 | 0.533 | 0.012 | 0.092 | 2 | 2 | 1 | 1 |
| $R_6$ | 0.496 | 0.488 | 0.022 | 0.097 | 2 | 2 | 1 | 1 |

*1) Real-world UCI data stream of chunks setup*

The first data stream we generated (Pen-Digits Data Stream) was created using the "Pen-Based Recognition of Handwritten Digits Data Set" [23]. This dataset, once processed, contained 10 class labels and 16 attributes. Therefore, we generated 16 sets of artificial class labels. This dataset was divided up into 10 data chunks with the records from each cluster label. This data stream contains a temporary drift that occurs at $t = 3$ and a sustaining concept drift from $t = 6$ onwards.

The last data steam we generated (Wine Quality Data Stream) was created using the "Wine Quality" dataset from UCI [23]. This dataset contained 7 class labels and 11 attributes. Therefore, we generated 11 sets of artificial class labels. This dataset was divided into 10 data chunks with the records from each cluster label. This data stream contains the same structure as the Pen-Digits Data Stream (a temporary drift that occurs at $t = 3$ and a sustaining concept drift from $t = 6$ onwards).

The overview of how both real-world datasets are divided into chunks is summarized in Table XI.

TABLE XI. REAL-WORLD DATA STREAM OF CHUNKS

| $t =$ | 1 | 2 | 3 | 4 | 5 | 6 | 7 | 8 | 9 | 10 |
|---|---|---|---|---|---|---|---|---|---|---|
| $|C^t|$ | 10 | 10 | 1 | 10 | 10 | 5 | 5 | 5 | 5 | 5 |

*2) Real-world UCI data stream of chunks results*

In our real-world scenario, we evaluated the clustering results of both SAIC and UIClust on the entropy, sum of squared errors, and runtime of both algorithms over both of our real-world UCI data streams. Because SAIC clusters based on class labels, and each dataset contains $n$ artificial class labels, we calculated the average entropy and SSE of all clusters, over each data stream for all artificial class sets. Our results showed the UIClust outperformed SAIC in all evaluation criteria. These results are summarized in Table XII.

TABLE XII. REAL-WORLD DATA STREAM RESULTS

| Evaluation Criteria (The lower the value, the better) | Pen-Digits Data stream | | Wine Quality Dataset | |
|---|---|---|---|---|
| | SAIC | UIClust | SAIC | UIClust |
| Entropy | 1.962 | **1.239** | 6.0275931 | **2.155766** |
| SSE | 77.604 | **14.877** | 3.3081654 | **0.61552** |
| Runtime (Seconds) | 26.15 | **6.22** | 3.69 | **1.97** |

## V. CONCLUSION AND FUTURE WORK

This paper presents an effective and efficient incremental clustering algorithm called UIClust that can cluster streams of data chunks. UIClust is designed to address two issues of clustering data streams. First, it addressed the memory limitation problem of clustering large streams of data chunks by summarizing cluster information into a series of key information. Secondly, the algorithm uses a novel parallel clustering component to handle temporary and sustaining concept drifts successfully. We compare the performance of UIClust with a modern incremental clustering algorithm called SAIC [16]. In our experiments, we used synthetic and real-world datasets (available on UCI ML repository [23]). In our synthetic dataset experiments, we generate four synthetic data streams in three different scenarios. In the first synthetic testing scenario, we evaluate the performance of UIClust with SAIC using the metrics entropy and the number of clusters discovered in the first data stream (SDWCD) to see how both algorithms handle concept drifts. The results indicated the performance of UIClust is the same as SAIC in terms of entropy. However, only UIClust discover the correct number of clusters in the data chunks. In the second synthetic scenario, we compare SAIC and UIClust's centroid values at the end of clustering the second synthetic data stream (SDCCL) with the data stream's clusters' TCV. The results show that UIClust manages to capture the TCV of each cluster while SAIC does not. This success is attributed to UIClust's novel parallel clustering component. For the last synthetic scenario, we compare SAIC and UIClust's runtimes on two large synthetic data streams. It was discovered that UIClust performed on average 24% quicker than SAIC. UIClust's ability to incrementally cluster the dataset quicker is due to only requiring a single pass when there is no concept drift. Finally, in our real-world dataset experiments, we evaluate the performance of UIClust and SAIC on two data streams generated from the UCI Pen-Digits and Wine Quality datasets. The experiment is designed to evaluate both algorithms' ability to handle data streams consisting of real-world data with the presence of a temporary and sustaining concept drift. The experimental results indicate the superiority of UIClust over SAIC in terms of entropy, SSE, and runtime.